\documentclass[11pt]{article}

\usepackage{acl}

\usepackage{times}
\usepackage{latexsym}
\usepackage{combelow}
\usepackage{xcolor}
\usepackage{graphicx}
\usepackage{pifont}
\usepackage{adjustbox}
\usepackage{booktabs}
\usepackage{multicol}
\usepackage{multirow}
\usepackage{floatrow}
\usepackage{svg}

\definecolor{tokencolor}{HTML}{BB3E00}
\definecolor{charcolor}{HTML}{F7AD45}

\usepackage{fullpage}
\usepackage{tikz}

\usepackage{tcolorbox}
\tcbuselibrary{minted,breakable,xparse,skins}

\usepackage{algorithm}
\usepackage{hyperref}
\usepackage{algpseudocode}
\usepackage{tikz-cd}

\usepackage{amssymb,amsmath,amsthm}

\usepackage[T1]{fontenc}
\usepackage[utf8]{inputenc}
\usepackage{microtype}

\theoremstyle{definition}
\newtheorem{definition}{Definition}[section]
\newtheorem{corrolary}{Corollary}[section]

\title{Training with Subword Variants Delays Overfitting in Repeated-Data Language Modeling}

\author{Adrian Cosma$^1$, Stefan Ruseti$^2$, Emilian Radoi$^2$, Mihai Dascalu$^2$\\
    {$^1$Dalle Molle Institute for Artificial Intelligence Research (IDSIA)} \\
    {$^2$National University of Science and Technology POLITEHNICA Bucharest (UNSTPB)} \\
    {\small\texttt{adrian.cosma@idsia.ch}}, 
    {\small\texttt{\{stefan.ruseti, emilian.radoi, mihai.dascalu\}@upb.ro}}}

\begin{document}
\maketitle
\begin{abstract}
Subword tokenization introduces a computational layer in language models, where many distinct token sequences decode to the same surface form while preserving meaning, yet induce different internal computations. Despite this non-uniqueness, language models are typically trained using a single canonical deterministic tokenization. In this work, we use non-canonical subword variants as a strictly meaning-preserving form of data augmentation. We introduce a lightweight training architecture that conditions canonical next-token prediction on sampled subword variants via an auxiliary causal encoder and block-causal cross-attention, without modifying the training objective or token interface. In data-constrained pretraining, augmentations with subword variants consistently delay overfitting under repeated data exposure and reduce held-out next-token loss across evaluation datasets. In multilingual fine-tuning, we find that the effectiveness of training with auxiliary subwords depends on tokenizer quality: gains are strongest when canonical tokens are highly compressed and diminish when the tokenizer already over-fragments the input. 
\end{abstract}

\section{Introduction}
\label{sec:intro}
The relationship between form and meaning is not arbitrary. The standard hierarchy of linguistic analysis (from phonology and morphology to semantics and pragmatics) describes increasing levels of abstraction from surface form to communicative intent \cite{jm3}. Classical linguistic theory places words at a privileged level in this hierarchy: lexical items mediate between form and meaning, while lower-level variations (e.g., phonetic realizations) are typically treated as semantically irrelevant. Modern Large Language Models (LLMs), however, introduce an additional computational layer that is largely invisible to human linguistic analysis: subword tokenization \cite{kudo-richardson-2018-sentencepiece,shibata1999byte}.

Subword tokenization algorithms such as Byte Pair Encoding (BPE) \cite{shibata1999byte} segment text into discrete units that are neither pure characters nor stable lexical items. These tokens sit at an intermediate level between characters, morphemes, and words, and are defined not by linguistic principles but by compression and frequency statistics \cite{shibata1999byte}. In practice, LLMs are trained and evaluated almost exclusively on deterministic canonical tokenizations, even though alternative segmentations of the same word are equally valid from a decoding perspective.

The token is the fundamental unit of computation in contemporary LLMs: it determines sequence length, attention patterns, and, ultimately, the model's internal activation dynamics. However, prior work has shown that training LLMs with BPE tokenizations is suboptimal \cite{bostrom2020byte}, and that finding an optimal tokenization (in terms of maximizing text compression) is computationally infeasible \cite{whittington2025tokenisation}. As a result, tokenization is typically treated as a fixed preprocessing step rather than an object of learning or perturbation.

There is conflicting evidence on whether LLMs are actually robust to tokenizations. For example,  \citet{zheng2025brokentokenslanguagemodel} showed that LLMs are robust to non-canonical tokenizations of the input, but a plethora of other works show that LLMs are severely bottlenecked by the underlying tokenization in multilingual settings \cite{unfairness-tokenizer,tokenization-meaning,wang2024tokenization}. LLMs are blind to the character composition of words \cite{shin2024large} and require a significant amount of data to overcome this limitation \cite{cosma-etal-2025-strawberry}. 

A natural response to tokenization sensitivity is to modify or replace the tokenizer itself. However, training better tokenizers or abandoning tokenization altogether often incurs substantial computational cost, architectural complexity, or technical debt \cite{pepe2024taxonomy}. In this work, we take a different approach: we treat the tokenizer as given and instead ask whether models can be trained to become invariant to tokenization choices and, by extension, to extract more signal out of the training dataset.

We use non-canonical subword segmentations\footnote{That is, different subword segmentations of the same underlying word that decode to the identical surface string.}, which prior work has studied under subword regularization \cite{kudo2018subword}, as auxiliary inputs aligned to the canonical token stream. While subword segmentations differ at the level of internal computation, producing distinct token sequences and activation patterns, they preserve the exact lexical identity and semantics of the text. This makes them a particularly attractive form of data augmentation as they alter the model’s internal representation without changing meaning.

Building on this observation, we propose training LLMs with both canonical deterministic tokenization and sampled non-canonical subword segmentations of the same input. We implement this using a lightweight auxiliary causal encoder that processes non-canonical subwords and injects their representations into the main decoder via block-causal cross-attention. Crucially, the standard next-token prediction objective and canonical token interface are preserved, avoiding ambiguity in the training signal. From a data augmentation perspective, these non-canonical subwords occupy a unique position. Unlike rephrasings or synonym replacements \cite{marivate2020improving}, which operate at the lexical or semantic level and often fail to preserve precise meaning \cite{lyons1995linguistic}, subword augmentations act at a lower computational abstraction level while preserving exact meaning. They therefore satisfy a strict notion of label preservation appropriate for language model pretraining, while being computationally cheap and compatible with other augmentation or regularization techniques.

Our contributions are as follows:

\begin{enumerate}
    \item We revisit alternative valid subword segmentations as a form of string-preserving subword regularization, and study their use as auxiliary aligned context for causal language modeling. To accommodate this form of data augmentation, we propose a simple, modular architectural modification to the standard transformer that enables training LLMs with auxiliary subwords without altering the standard language modeling objective.
    
    \item We empirically show that pretraining with auxiliary subwords delays overfitting and improves data efficiency by increasing variation in the training signal, while remaining compatible with other activation-level perturbation methods.
    
    \item We establish that the effectiveness of our approach depends on tokenizer quality in multilingual fine-tuning scenarios: performance improvements are greater when the input text is less fragmented by the tokenizer.
\end{enumerate}

\section{Related Work}
\label{sec:related}
\subsection{Impact of Tokenization in LLM Training}

Subword tokenization \cite{shibata1999byte,provilkov2020bpe,zouhar2023formal,zouhar2023tokenization} was developed to address the issue of out-of-vocabulary words when considering individual words as fundamental units of computation. Indeed, defining a fixed vocabulary size is bound to introduce out-of-vocabulary words, either due to misspellings, but also because the morphological derivational processes make the lexicon unbounded \cite{aronoff2022morphology}. Using raw bytes as input units is undesirable as it is computationally inefficient and removes necessary inductive biases that enable models to generalize \cite{rajaraman2024toward}. 

However, multiple studies have shown that subword tokenization impacts the multilingual performance of language models \cite{limisiewicz2023tokenization} and introduces unfairness across languages \cite{unfairness-tokenizer} since most tokenizers are trained on English-centric corpora. Subsequently, \citet{jabbar2023morphpiece} found that designing a linguistically informed tokenizer improves performance. The root cause of this effect is that language models are blind to the character composition of their tokens \cite{shin2024large,cosma-etal-2025-strawberry}. This problem cannot be readily solved by simply scaling the training data \cite{cosma-etal-2025-strawberry} or by training the tokenizer on more data \cite{goldman2024unpacking}. For example, \citet{reddy2025much} found diminishing returns for increasing tokenizer training data, with a more pronounced effect on languages with extensive inflectional and derivational morphology (e.g., Finnish, Hungarian, and Romanian, \textit{inter alia}).

Other approaches attempt to circumvent the need for tokenization by designing neural architectures that operate directly on bytes \cite{slagle2024spacebyte,ahia2024magnet,deiseroth-etal-2024-free,nawrot2023efficient,pagnoni2025byte}, but incur additional computational costs and technical debt \cite{pepe2024taxonomy} by deviating from known and relatively understood architectural design choices of the modern Transformer \cite{vaswani2017attention}. 

We depart from these approaches and instead adopt a data-driven approach to increase the model's perceptual range: the tokenizer is a given, and we encourage robustness to alternative tokenizations by exposing the model to non-canonical tokenizations of the same underlying text as a form of data augmentation.

\subsection{Data Augmentation for Text}
\label{sec:related-augs}

Data augmentation is widely used in other domains, such as Computer Vision \cite{yang2022image,image-augs-survey}, enabling much longer training runs, up to hundreds of epochs, especially when the dataset size is limited \cite{geiping2022much,villalobos2022will}. Data augmentation induces data-driven invariances in the model \cite{geiping2022much,balestriero2022aug,chen2020group} and increases robustness to OOD data \cite{geiping2022much}. Most such augmentations for Computer Vision are cheaply generated using programmatic operations (e.g., random flipping, cropping, and color jitter). However, cheap data augmentations for text data have remained elusive. \citet{chai2025text} surveyed approaches to text data augmentation for large language models, but most require substantial computational overhead. For example, the Kimi-K2 \cite{team2025kimi} model employed large-scale rephrasing with another pretrained model during the construction of the pretraining dataset to increase the count of high-quality tokens. 

In general, effective text augmentations have been limited to back-translations or synonym replacements \cite{marivate2020improving}. It is unclear, however, if such augmentations preserve the precise meaning and syntax of the original text -- augmentations should be "label preserving" \cite{taylor2018improving}. Nevertheless, in the case of language modeling, augmentation should be "meaning preserving". Back-translations are bottlenecked by translation model performance and lexical-semantic gaps between languages \cite{pimentel2020speakers}. Furthermore, there are a few exact synonyms across languages \cite{edmonds2002near,lyons1995linguistic,quine2013word}, suggesting that augmentations that rely on synonym replacements subtly change the meaning of the original text.

Prior work has explored alternative subword segmentations through subword regularization \cite{kudo2018subword}, BPE-dropout \cite{chai-etal-2024-tokenization}, and MaxMatch-style dropout \cite{hiraoka2022maxmatch}. These methods are most relevant to our setting because they modify the token sequence while preserving the decoded string. Our work differs not in observing that alternative segmentations exist, but in how they are used: rather than replacing the model input with a sampled segmentation, we keep the canonical causal LM stream and use non-canonical segmentations as aligned auxiliary context. This preserves the standard next-token label space and avoids redefining the language-modeling objective.

Further, the work from \citet{prabhudesai2025diffusion} suggested that the training process of text diffusion models acts as a form of implicit data augmentation, which alleviates the diminishing returns of repeating data and enables longer training \cite{muennighoff2025scalingdataconstrainedlanguagemodels}.

\section{Method}
\label{sec:method}
\begin{figure*}[hbt!]
    \centering
    \includesvg[width=1.0\linewidth]{images/strawberry-v2-diagram.drawio.svg}
    \caption{Diagram of our architecture. The main causal decoder operates on canonical BPE tokens, while a lightweight single-block causal encoder consumes sampled subword variants (e.g., \texttt{dinosaur} $\rightarrow$ \texttt{dino}~+~\texttt{saur} | \texttt{d}~+~\texttt{inosaur}) and injects them into the main trunk via a block-causal cross-attention operation.}
    \label{fig:diagram}
\end{figure*}

\subsection{Augmentations across Levels of Linguistic Abstraction}

As discussed in Section \ref{sec:related-augs}, augmentations in supervised tasks are perturbations of the input that preserve class labels. In Computer Vision, it is clear that pixel transformations (unless severe) do not change the class label. However, each linguistic text unit serves a definite purpose in communicating intent and in defining semantics, calling into question the efficacy of current augmentation techniques. Intuitively, augmentations that have the highest impact on the model should be at higher levels of abstractions, to make the model invariant to surface forms. However, altering surface forms at a higher level of abstraction than words does change the underlying meaning of the text. In language model pretraining, supervision is implicit, and, in this case, augmentations should precisely preserve the meaning of the underlying unit, even if the surface form structure changes.

To formalize this intuition, we consider the different levels of linguistic abstractions \cite{jm3}. Let $\mathcal{X}_0 \subset \mathcal{X}_1 \subset \cdots \subset \mathcal{X}_L$ denote progressively higher linguistic levels (e.g., morphology $\rightarrow$ lexicality $\rightarrow$ semantics). Each level $\ell$ has an associated projection map $\pi_{\ell+1}: \mathcal{X}_\ell \rightarrow \mathcal{X}_{\ell+1}$ that identifies the linguistic unit at the next abstraction level.

\begin{definition}[Text Augmentation]
\label{def:aug}
An augmentation at level $\ell$ is a transformation $a : \mathcal{X}_\ell \rightarrow \mathcal{X}_\ell$
that preserves the \textit{supervisory signal} defined at level $\ell+1$. Formally, if $x \in \mathcal{X}_{\ell}$, then $\pi_{l+1}(x) = \pi_{l+1}(a(x))$.
\end{definition}

Applying the augmentation does not change the representation at the higher level. Given the limitations of existing methods, we claim that current text augmentations that operate on levels beyond the lexical level, such as rephrasings and synonym replacements \cite{marivate2020improving}, violate Definition \ref{def:aug}. To further support our claim from a linguistic perspective, we consider the classical distinctions in lexical semantics, namely that the underlying assumption of synonym replacements is absolute synonymy, which is known to be exceedingly rare except in technical registers \cite{lyons1995linguistic}; in a given context, each word slightly changes the meaning and intent of the text. Furthermore, in practice, low-resource languages often lack comprehensive lexical databases (e.g., WordNet \cite{miller-1994-wordnet}) and exhibit poor back-translation quality, influencing downstream tasks \cite{agrawal2024translation}.

\subsection{Subword Variants as Meaning-Preserving Text Augmentations}

In practice, language models use subword tokenization \cite{shibata1999byte}, which introduces a computational layer between the orthographic surface and the lexical level. A single word $w$ may admit multiple valid segmentations under a stochastic tokenizer. These segmentations differ in their subword decomposition but correspond to the same lexical item and the same meaning. This motivates the following definitions.

\begin{definition}[Canonical and Non-Canonical Tokenizations]
\label{def:tok}
Let $w \in \Sigma^*$ be a string span. Let $\mathcal{T}:\Sigma^* \rightarrow V^*$ denote the deterministic tokenization
procedure associated with a fixed pretrained tokenizer, and $\hat{\mathcal{T}}$ a stochastic version of the same tokenizer which randomly splits the word into several tokens from the token vocabulary $V$. Then:

\begin{itemize}
    \item The canonical tokenization is the deterministic mapping $x = \mathcal{T}(w)$ such that $\mathcal{T}^{-1}(x) = w$.
    \item A non-canonical tokenization is any $s = \hat{\mathcal{T}}(w)$, consisting of tokens $(s_1,\dots,s_k) \in V$ such that $\mathcal{T}^{-1}(s) = w$. 
\end{itemize}
\end{definition}

For BPE tokenizers, canonical tokenization is determined by the learned merge table; for WordPiece tokenizers \cite{devlin2019bert}, it is commonly implemented through greedy longest-prefix matching. Our method does not rely on either property, only on the existence of a deterministic canonical tokenization and alternative valid segmentations that decode to the same string.

For example, the word \textit{"dinosaur"}, using a pretrained GPT-2 tokenizer, is canonically tokenized as \{\textit{d}$_{67}$, \textit{inosaur}$_{21317}$\}, whereas a non-canonical tokenization would be: \{\textit{d}$_{67}$, \textit{ino}$_{2879}$, \textit{s}$_{82}$, \textit{aur}$_{2899}$\}, which is decoded into the same string. All tokens are part of the same tokenizer vocabulary.

\begin{definition}[Subword Variants]
    \label{def:homo}
    Two sequences $s$ and $s'$ are said to be \textit{subword variants} if they arise from different valid segmentations of the same word. That is, $\mathcal{T}^{-1}(s) = \mathcal{T}^{-1}(s')$.
\end{definition}

\begin{corrolary}
    Stochastic tokenizations are a form of text augmentation.
\end{corrolary}
\begin{proof}
    Considering $\pi_{\ell+1} = \mathcal{T}^{-1}$, the transformation from the token level to the lexical level $\mathcal{X}_{\text{tok}} \rightarrow \mathcal{X_{\text{lex}}}$, from Definitions \ref{def:aug}, \ref{def:tok} and \ref{def:homo} it is clear that $\hat{\mathcal{T}}$ is an augmentation.
\end{proof}


Although non-canonical tokenizations modify the internal computational representation (changing the token sequence seen by the model), they preserve the \textit{exact} word identity and thus, meaning of the text. Because tokenization determines the discrete units processed by an LLM, two segmentations of the same word produce different activation patterns, even though they correspond to the same lexical item and the same semantics. 
\subsection{Training with \textit{subword variants}}
\label{subsec:subword-variants-training}

Given a large corpus $\mathcal{D}$, the next-token-prediction modelling objective ubiquitously used in language models implies sampling a random text $x \sim \mathcal{D}$ and applying the canonical tokenizer to obtain $s = \mathcal{T}(x) = (s_1, s_2, \dots,s_K)$, the individual canonical tokens of $x$, having the sequence length $K$. The objective is to model $P(s_{t+1} | s_1, s_2, \dots,s_t)$, $t<K$. 

Consider the subword variants of each canonical token: $\hat{s} = (\hat{\mathcal{T}}(s_1), \dots, \hat{\mathcal{T}}(s_K))$, with $\hat{\mathcal{T}}(s_i) = (\hat{s}_{i1}, \dots, \hat{s}_{iL_i})$, having the larger context size $K \leq K' = \sum_{i=0}^{k} L_i$. Training with subword variants implies that the objective is to model $P(s_{t+1} | s_{1}, \dots,s_{t}; \hat{s}_1, \dots\hat{s}_{tL_t})$. That is, estimating the probability of the next canonical token, while accounting for all previous canonical tokens and their sampled subword variants. We further describe the architectural details that best accommodate this setup.

\paragraph{Causal encoder and cross-attention.}

We decided to adopt an encoder-decoder style design \cite{cosma-etal-2025-strawberry,provilkov2020bpe} tailored to causal language modeling, as shown in Figure \ref{fig:diagram}. In Appendix \ref{sec:alt-designs} we briefly discuss alternative designs that we considered, but discarded. The main trunk is a standard causal decoder operating on canonical tokens $(s_1,\dots,s_k)$. In parallel, a significantly smaller \textit{causal encoder} branch processes the subword sequence $\hat{s}$. The augmentation branch consists of a single Transformer block, which encodes $\hat{s}$ under a block causal mask $M$. This causal mask is constructed as follows:

\begin{equation}
    \mathbf{M} =
    \begin{pmatrix}
        \mathbf{1}_{L_1\times L_1} & 0 & \cdots & 0 \\
        \mathbf{1}_{L_2\times L_1} & \mathbf{1}_{L_2\times L_2} & \cdots & 0 \\
        \vdots & \vdots & \ddots & \vdots \\
        \mathbf{1}_{L_K\times L_1} &
        \mathbf{1}_{L_K\times L_2} & \cdots &
        \mathbf{1}_{L_K\times L_K}
    \end{pmatrix},
    \label{eq:block-causal-mask}
\end{equation}
where $K$ is the number of canonical tokens and each block $\mathbf{1}_{L_i\times L_j}$ denotes an all-ones matrix of shape $L_i \times L_j$, with $L_i$ being the length of the subword variant $i$. 

This mask ensures that all subtokens from a token $s_t$ can attend only to subtokens of tokens $1,\dots,t$ (including themselves), preserving causality at the canonical token level.

The encoded subword variants representations are injected into the first block of the main trunk via cross-attention using a block-causal cross-attention mask $M_{\text{cross}}$.  This mask is constructed as:
\begin{equation}
    \mathbf{M}_\text{cross} =
    \begin{pmatrix}
        \mathbf{1}_{L_1} & 0 & \cdots & 0 \\
        \mathbf{1}_{L_1} & \mathbf{1}_{L_2} & \cdots & 0 \\
        \vdots & \vdots & \ddots & \vdots \\
        \mathbf{1}_{L_1} & \mathbf{1}_{L_2} & \cdots & \mathbf{1}_{L_K}
    \end{pmatrix},
    \label{eq:block-cross-mask}
\end{equation}
where $\text{\textbf{1}}_{L_i}$ is a shorthand for $\text{\textbf{1}}_{1 \times L_i}$. Row $t$ specifies which subtokens are visible to the canonical token $s_t$ during cross-attention. 

This yields a word-level causal structure: $s_t$ can attend to the subtokens of tokens $1,\dots,t$, but not to those of future tokens. The tokenizer typically splits the canonical tokens into single characters/bytes; having the subtoken sequence consisting mostly of single characters increases the sequence length, which affects computational speed, but also prevents proper training of longer subtokens that are more likely to be canonical tokens in other pieces of text. Consequently, subword variants are constructed by randomly sampling from the top 5 longest subtokens by character count, excluding the canonical token, thereby balancing granularity and computational cost. If the canonical token cannot be subdivided, we use it as such. At inference time, we use the same procedure. 

Since attention is permutation-invariant over its input set \cite{vaswani2017attention,lee2019set}, we explicitly encode token order using learned positional embeddings: order within each subword variant sequence is encoded using \textit{intra-token position embeddings}, and the correspondence between canonical tokens $s_i$ and their subword variant sequences $(\hat{s_{ij}})_{j\leq L_i}$ is encoded using \textit{inter-token position embeddings}. This design is inspired by Abacus position embeddings \cite{mcleish2024abacus} that explicitly encode the position of individual digits inside a number. RoPE \cite{rope} cannot be used without some modifications for the subword variant branch, since it will induce a mismatch between subword variants positions and canonical token positions. This problem has been addressed, for example, by considering fractional position embeddings \cite{harzig2022synchronized,gomez2024depression} that align the sampling rates of videos and associated audio. For a fair comparison, both the standard transformer used in our experiments and our variant use learned position embeddings, similar to GPT-2 \cite{radford2019language}. 

Note that such subword variants augmentations are \textit{stack-compatible} with other data- or model-level augmentations. Rephrasing, dropout, or other regularization schemes operate orthogonally to our subword variant branch; they can be applied in combination with our method without changing the model interface or loss definition. 

This design is suggestive of the cross-resolution fusion in Computer Vision architectures such as CrossViT \cite{chen2021crossvit}, where high-resolution features enrich coarse representations. This high-level design pattern has been used in NLP \cite{he2024hdt,chalkidis2022exploration,wu-etal-2021-hi,cosma-etal-2025-strawberry}, but it is far more common in Computer Vision \cite{wang2021pyramid,liu2021swin,chu2021twins,xu2021co,wang2023crossformer,chen2021regionvit}. Furthermore, this can also be seen as an attention adapter \cite{han2024parameter}, which can presumably be used to retrofit existing pretrained models to enhance their sub-token perception capabilities.

\section{Experimental Setup}
\label{sec:exp}
\begin{figure*}[hbt!]
    \centering
    \includesvg[width=1.0\linewidth]{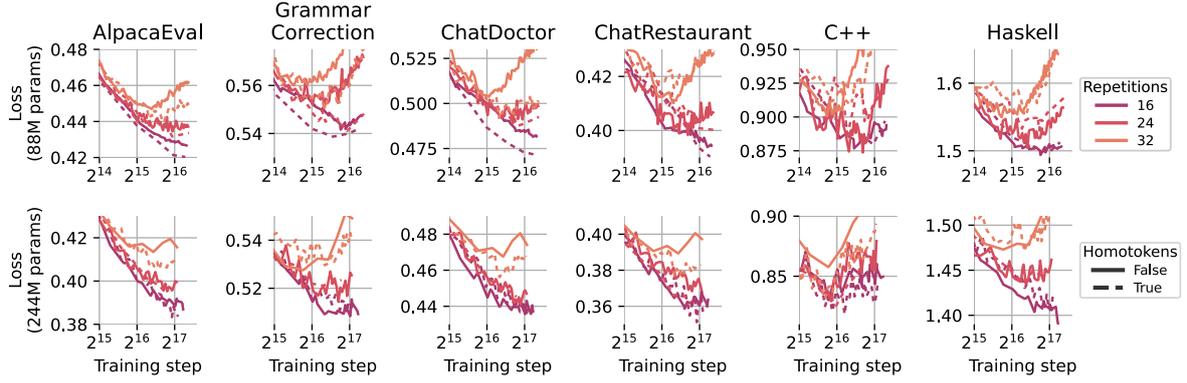}
    \caption{Evaluations on various chat datasets across pretraining duration for two model sizes: $\mu=1$ - \textbf{top} and $\mu=2$ - \textbf{bottom}. We show training runs with the number of data repetitions $R \in \{16, 24, 32, 64\}$ for clarity. Training with subword variants consistently delays overfitting in these scenarios.}
    \label{fig:evals}
\end{figure*}

We compared our architectural variant with a standard transformer decoder language model in data-constrained scenarios. Details about the training hyperparameters and evaluation protocol can be found in Appendices \ref{sec:hyperparams} and \ref{sec:eval}.

\noindent \textbf{Pretraining configuration.}
For pretraining, we used a subset of FineWeb-Edu \cite{penedo2024fineweb}. We follow the setup by \citet{muennighoff2025scalingdataconstrainedlanguagemodels} to evaluate our model in data-constrained settings during pretraining. We increased the number of dataset repetitions $R$ while simultaneously sampling $1/R$ of the dataset to maintain Chinchilla-optimality \cite{hoffmann2022training} in terms of compute: the dataset size is approximately 20 tokens per model parameter, regardless of model size and number of repetitions. In particular,  we train on 4B tokens for $\mu=1$, and we train on 16B tokens for $\mu=2$. Unless specified, we used the GPT-2 \cite{radford2019language} tokenizer\footnote{\url{hf.co/openai-community/gpt2}, Accessed: 5 January 2026}. Since $\mu$P only addresses model size and not token horizon, we also adjust the learning rate according to the dataset size \cite{bjorck2024scaling}: $lr_{\mu} = lr_{\mu=1} \times (N_{params}(\mu) / N_{params}(\mu=1))^{-0.32}$. 

\noindent \textbf{Activation-level Perturbations.}
To compare with alternative activation-level augmentations, we pretrained separate models using attention dropout at 10\% and 20\% and Gaussian noise injected at the token level with a standard deviation of 0.1. We used these perturbations for both the vanilla transformer and our variant for $R = 32$ and $\mu = 1$ using the GPT-2 tokenizer.

\noindent \textbf{Multilingual Fine-tuning Configuration.}
A situation in which repeating data is common is instruction fine-tuning, and especially so in multilingual settings and when dealing with low-resource languages. Consequently, we pretrained each model variant on FineWeb-Edu using the GPT-2 tokenizer (vocabulary size $\sim$50k) and the aya-23 multilingual tokenizer (vocabulary size $\sim$255k) \cite{aryabumi2024aya} with one repetition of data and then finetuned for 32 epochs on a subset of 14 languages from the AYA dataset \cite{singh2024aya}, containing instructions in multiple languages. 

\subsection{Evaluation Protocol.} 
\label{sec:eval}
During pretraining, test loss is computed on a fixed held-out subset of FineWeb-Edu. We also evaluated on several instruction datasets from heterogeneous domains, targeting general knowledge through Alpaca Instructions \cite{alpaca}, lexical understanding through Grammar Correction\footnote{\url{hf.co/datasets/agentlans/grammar-correction}, Accessed: 5 January 2026}, code understanding through Haskell problems\footnote{\url{hf.co/datasets/finbarr/rlvr-code-data-haskell-edited}, Accessed: 5 January 2026}, and C++ problems\footnote{\url{hf.co/datasets/ReySajju742/synthetic-cpp}, Accessed: 5 January 2026} and two domain-specific chat datasets: ChatDoctor \cite{li2023chatdoctor} and ChatRestaurant\footnote{\url{hf.co/datasets/tctsung/chat_restaurant_recommendation}, Accessed: 5 January 2026}. Since the models are too small to generate meaningful text, we computed the average next-token prediction across concatenated question-answer pairs for each dataset.

\section{Results}
\label{sec:results}
\begin{figure*}
    \centering
    \includesvg[width=1.0\linewidth]{images/aya.svg}
    \caption{Performance across multilingual fine-tuning for $\mu = 1$ across 32 epochs. The model pretrained with the GPT-2 tokenizer (\textbf{top}) does not improve with auxiliary subword variants, whereas the variant trained with \textit{aya-23} tokenizer (\textbf{bottom}) and subword variants consistently outperforms the standard Transformer decoder when repeating data.}
    \label{fig:aya}
\end{figure*}

\noindent\textbf{Pretraining Behaviour.} In Figure \ref{fig:pretrain} we show evaluation curves for runs trained using different amounts of data repetitions, in a similar setup to \citet{muennighoff2025scalingdataconstrainedlanguagemodels}. We observe that when the data is repeated more than 16 times, training with subword variants offers clear advantages: the overfitting point is later in training, and the best loss is lower than that of the standard Transformer. For $R < 16$, training with subword variants does not affect final performance. For $\mu = 1$, the addition of subword variants seems somewhat detrimental at lower amounts of repetitions. Still, this effect is not held for $\mu=2$, where both models have essentially the same evaluation curves for $R < 24$. 

In Figure \ref{fig:evals}, we show the evolution of cross-domain LM evaluation losses during pretraining across several instruction-following and domain-specific chat datasets, for two model sizes and varying numbers of data repetitions. Across most benchmarks with sufficiently high repetition factors ($R \geq 16$), models trained with subword variants exhibit delayed overfitting compared to the standard Transformer baseline. While both models initially improve at a similar rate, the models augmented with subword variants maintain lower evaluation loss for a more extended period of training, with the gap becoming more pronounced as repetition increases. This effect is observed across heterogeneous domains, suggesting that the benefit is not task-specific but instead reflects a more general regularization effect. These results are consistent with the hypothesis that exposing the model to multiple meaning-preserving tokenizations 
enables it to continue extracting useful signal from repeated data instead of overfitting to a single canonical segmentation.

\begin{figure}[hbt!]
    \centering
    \includesvg[width=1.0\linewidth]{images/pretrain.svg}
    \caption{Pretraining curves for various numbers of data repetitions for an 88M parameter model \textbf{($\mu = 1$, left)} and a 244M parameter model \textbf{($\mu = 2$, right)}. Training with subword variants delays overfitting for $R > 16$.}
    \label{fig:pretrain}
\end{figure}

\begin{figure}[hbt!]
    \centering
    \includesvg[width=1.0\linewidth]{images/ad-gn.svg}
    \caption{Loss during pretraining for runs using attention dropout and token-level Gaussian noise as augmentations for $R = 32$ and $\mu =1$.}
    \label{fig:ad-gn}
\end{figure}
    
\begin{figure}[hbt!]
    \centering
    \includesvg[width=1.0\linewidth]{images/ad-gn-evals.svg}
    \caption{Loss across two benchmarks for runs using attention dropout (\textbf{top)} and token-level Gaussian noise (\textbf{bottom}) as augmentations for $R = 32$ and $\mu = 1$.}
    \label{fig:ad-gn-evals}
\end{figure}

\paragraph{Comparison with Activation-level Perturbations.}
In Figure \ref{fig:ad-gn}, we show the pretraining loss comparing our model with the standard Transformer under activation-level perturbations. Since our architecture is stack-compatible with other augmentations, we can also apply these perturbations to it, further increasing the amount of data augmentation. In both cases, training with subword variants markedly improves performance. In Figure \ref{fig:ad-gn-evals}, we show performance on the AlpacaEval and GrammarCorrection datasets across training; training with auxiliary subword variants benefits model generalization beyond simple noise injection.

\paragraph{Multilingual Fine-tuning.}
In Figure \ref{fig:aya}, we show the performance across multilingual instruction fine-tuning for models using the GPT-2 and aya-23 tokenizer. These results highlight the fact that training with subword variants is not a free lunch in all situations. We hypothesize that the greatest benefits of our approach stem from tokenizers that heavily compress the input sequence, yielding large canonical tokens. This is because if a tokenizer is poorly trained or used in out-of-distribution settings, the resulting canonical tokens are already fragmented and cannot be divided further, making our causal encoder to essentially process the same sequence as the main trunk. In our case, training with auxiliary subword variants and with the GPT-2 tokenizer, which is geared towards English, does not benefit multilinguality. In contrast, using the aya-23 tokenizer, which is specifically designed to handle multiple languages, shows definite benefits in this setup, obtaining better performance across training (see \ref{sec:tok-stats}, Table \ref{tab:tokenizer-stats} - for example, Finnish and Zulu show substantially higher tokens/word ratios). This shows that our method should not be used naively in multilingual settings, expecting improvements without addressing the underlying over-fragmentation issues with the tokenizer. Our method is most appropriately used to introduce more variation in the training data and is synergistic with current trends in scaling tokenizer vocabulary size \cite{huang2025overtokenized}.

\section{Conclusions}
\label{sec:conc}
We studied the effect of auxiliary subword regularization under repeated-data training. By exposing language models to multiple valid segmentations of the same lexical items, without altering the canonical next-token prediction objective, we provide evidence that aligned non-canonical segmentations can regularize repeated-data training; whether this corresponds to representation-level tokenization invariance remains an open mechanistic question. Across data-constrained pretraining regimes, this approach consistently delays overfitting and reduces held-out next-token loss across evaluation datasets, particularly under repeated data exposure, while remaining compatible with standard architectures and existing regularization techniques. Our results suggest that tokenization should not be treated as a fixed preprocessing artifact, but as a source of structured variation that can be leveraged during training. More broadly, leveraging subword variants offers a practical path toward more robust and data-efficient language model training without abandoning established tokenization pipelines.

\section*{Limitations}
\label{sec:lim}
As this study is intended as a proof of concept, its primary limitation is its limited scale due to hardware constraints. All experiments are conducted using relatively small language models and data-constrained training regimes. While this setup is well-suited for isolating overfitting behavior, we cannot claim that augmentations based on subword variants would produce comparable gains in large-scale ($>1B$ parameter models) pretraining settings. Our reported evaluations use the auxiliary segmentation branch at both training and evaluation time. Thus, the method should be understood as auxiliary tokenization conditioning rather than train-only data augmentation. We do not claim that the main decoder fully internalizes the augmentation when the auxiliary branch is removed; evaluating train-only use is left for future work. The present contribution shows evidence for the value of aligned auxiliary segmentations in repeated-data training.

\section*{Acknowledgments}
\label{sec:ack}
This work was supported by UBS Switzerland AG and its affiliates, and by the project “Romanian Hub for Artificial Intelligence - HRIA”, Smart Growth, Digitization and Financial Instruments Program, 2021-2027, MySMIS no. 351416.

\bibliography{refs}

@inproceedings{hiraoka2022maxmatch,
  title={MaxMatch-dropout: Subword regularization for WordPiece},
  author={Hiraoka, Tatsuya},
  booktitle={Proceedings of the 29th International Conference on Computational Linguistics},
  pages={4864--4872},
  year={2022}
}

@inproceedings{devlin2019bert,
  title={Bert: Pre-training of deep bidirectional transformers for language understanding},
  author={Devlin, Jacob and Chang, Ming-Wei and Lee, Kenton and Toutanova, Kristina},
  booktitle={Proceedings of the 2019 conference of the North American chapter of the association for computational linguistics: human language technologies, volume 1 (long and short papers)},
  pages={4171--4186},
  year={2019}
}

@inproceedings{kudo-richardson-2018-sentencepiece,
    title = "{S}entence{P}iece: A simple and language independent subword tokenizer and detokenizer for Neural Text Processing",
    author = "Kudo, Taku  and
      Richardson, John",
    editor = "Blanco, Eduardo  and
      Lu, Wei",
    booktitle = "Proceedings of the 2018 Conference on Empirical Methods in Natural Language Processing: System Demonstrations",
    month = nov,
    year = "2018",
    address = "Brussels, Belgium",
    publisher = "Association for Computational Linguistics",
    url = "https://aclanthology.org/D18-2012/",
    doi = "10.18653/v1/D18-2012",
    pages = "66--71"
}

@article{bostrom2020byte,
  title={Byte pair encoding is suboptimal for language model pretraining},
  author={Bostrom, Kaj and Durrett, Greg},
  journal={arXiv preprint arXiv:2004.03720},
  year={2020}
}

@article{shibata1999byte,
  title={Byte pair encoding: A text compression scheme that accelerates pattern matching},
  author={Shibata, Yusuxke and Kida, Takuya and Fukamachi, Shuichi and Takeda, Masayuki and Shinohara, Ayumi and Shinohara, Takeshi and Arikawa, Setsuo},
  year={1999},
  publisher={Technical Report DOI-TR-161, Department of Informatics, Kyushu University}
}

@Book{jm3,
  author =       "Daniel Jurafsky and James H. Martin",
  title =        "Speech and Language Processing: An Introduction to Natural Language Processing, 
  		  Computational Linguistics, and Speech Recognition,
		   with Language Models",
  year =         "2025",
  url = {https://web.stanford.edu/~jurafsky/slp3/},
  note = "Online manuscript released August 24, 2025",
  edition =         "3rd",
  }

@book{aronoff2022morphology,
  title={What is morphology?},
  author={Aronoff, Mark and Fudeman, Kirsten},
  year={2022},
  publisher={John Wiley \& Sons}
}

@article{rajaraman2024toward,
	title        = {Toward a theory of tokenization in llms},
	author       = {Rajaraman, Nived and Jiao, Jiantao and Ramchandran, Kannan},
	year         = 2024,
	journal      = {arXiv preprint arXiv:2404.08335}
}

@article{goldman2024unpacking,
  title={Unpacking tokenization: Evaluating text compression and its correlation with model performance},
  author={Goldman, Omer and Caciularu, Avi and Eyal, Matan and Cao, Kris and Szpektor, Idan and Tsarfaty, Reut},
  journal={arXiv preprint arXiv:2403.06265},
  year={2024}
}

@article{rope,
author = {Su, Jianlin and Ahmed, Murtadha and Lu, Yu and Pan, Shengfeng and Bo, Wen and Liu, Yunfeng},
title = {RoFormer: Enhanced transformer with Rotary Position Embedding},
year = {2024},
issue_date = {Feb 2024},
publisher = {Elsevier Science Publishers B. V.},
address = {NLD},
volume = {568},
number = {C},
issn = {0925-2312},
url = {https://doi.org/10.1016/j.neucom.2023.127063},
doi = {10.1016/j.neucom.2023.127063},
journal = {Neurocomput.},
month = feb,
numpages = {12}
}

@misc{huang2025overtokenized,
	title        = {Over-Tokenized Transformer: Vocabulary is Generally Worth Scaling},
	author       = {Hongzhi Huang and Defa Zhu and Banggu Wu and Yutao Zeng and Ya Wang and Qiyang Min and Xun Zhou},
	year         = 2025,
	url          = {https://arxiv.org/abs/2501.16975},
	eprint       = {2501.16975},
	archiveprefix = {arXiv},
	primaryclass = {cs.CL}
}

@inproceedings{slagle2024spacebyte,
	title        = {SpaceByte: Towards Deleting Tokenization from Large Language Modeling},
	author       = {Kevin Slagle},
	year         = 2024,
	booktitle    = {The Thirty-eighth Annual Conference on Neural Information Processing Systems},
	url          = {https://openreview.net/forum?id=KEe4IUp20I}
}

@article{kingma2014adam,
  title={Adam: A method for stochastic optimization},
  author={Kingma, Diederik P and Ba, Jimmy},
  journal={arXiv preprint arXiv:1412.6980},
  year={2014}
}

@article{penedo2024fineweb,
  title={The fineweb datasets: Decanting the web for the finest text data at scale},
  author={Penedo, Guilherme and Kydl{\'\i}{\v{c}}ek, Hynek and Lozhkov, Anton and Mitchell, Margaret and Raffel, Colin A and Von Werra, Leandro and Wolf, Thomas and others},
  journal={Advances in Neural Information Processing Systems},
  volume={37},
  pages={30811--30849},
  year={2024}
}

@misc{yang2022tensorprogramsvtuning,
      title={Tensor Programs V: Tuning Large Neural Networks via Zero-Shot Hyperparameter Transfer}, 
      author={Greg Yang and Edward J. Hu and Igor Babuschkin and Szymon Sidor and Xiaodong Liu and David Farhi and Nick Ryder and Jakub Pachocki and Weizhu Chen and Jianfeng Gao},
      year={2022},
      eprint={2203.03466},
      archivePrefix={arXiv},
      primaryClass={cs.LG},
      url={https://arxiv.org/abs/2203.03466}, 
}

@article{loshchilov2016sgdr,
  title={Sgdr: Stochastic gradient descent with warm restarts},
  author={Loshchilov, Ilya and Hutter, Frank},
  journal={arXiv preprint arXiv:1608.03983},
  year={2016}
}

@article{geiping2022much,
  title={How much data are augmentations worth? an investigation into scaling laws, invariance, and implicit regularization},
  author={Geiping, Jonas and Goldblum, Micah and Somepalli, Gowthami and Shwartz-Ziv, Ravid and Goldstein, Tom and Wilson, Andrew Gordon},
  journal={arXiv preprint arXiv:2210.06441},
  year={2022}
}

@inproceedings{taylor2018improving,
  title={Improving deep learning with generic data augmentation},
  author={Taylor, Luke and Nitschke, Geoff},
  booktitle={2018 IEEE symposium series on computational intelligence (SSCI)},
  pages={1542--1547},
  year={2018},
  organization={IEEE}
}

@inproceedings{marivate2020improving,
  title={Improving short text classification through global augmentation methods},
  author={Marivate, Vukosi and Sefara, Tshephisho},
  booktitle={International Cross-Domain Conference for Machine Learning and Knowledge Extraction},
  pages={385--399},
  year={2020},
  organization={Springer}
}

@article{chai2025text,
  title={Text data augmentation for large language models: A comprehensive survey of methods, challenges, and opportunities},
  author={Chai, Yaping and Xie, Haoran and Qin, Joe S},
  journal={arXiv preprint arXiv:2501.18845},
  year={2025}
}

@article{chen2020group,
  title={A group-theoretic framework for data augmentation},
  author={Chen, Shuxiao and Dobriban, Edgar and Lee, Jane H},
  journal={Journal of Machine Learning Research},
  volume={21},
  number={245},
  pages={1--71},
  year={2020}
}

@article{ahia2024magnet,
  title={Magnet: Improving the multilingual fairness of language models with adaptive gradient-based tokenization},
  author={Ahia, Orevaoghene and Kumar, Sachin and Gonen, Hila and Hofmann, Valentin and Limisiewicz, Tomasz and Tsvetkov, Yulia and Smith, Noah A},
  journal={Advances in Neural Information Processing Systems},
  volume={37},
  pages={47790--47814},
  year={2024}
}

@article{zhang2019root,
  title={Root mean square layer normalization},
  author={Zhang, Biao and Sennrich, Rico},
  journal={Advances in neural information processing systems},
  volume={32},
  year={2019}
}

@book{lyons1995linguistic, 
    place={Cambridge}, 
    title={Linguistic Semantics: An Introduction}, 
    publisher={Cambridge University Press}, 
    author={Lyons, John},
    year={1995}
}

@book{quine2013word,
  title={Word and object},
  author={Quine, Willard Van Orman},
  year={2013},
  publisher={MIT press}
}

@article{bjorck2024scaling,
  title={Scaling optimal LR across token horizons},
  author={Bjorck, Johan and Benhaim, Alon and Chaudhary, Vishrav and Wei, Furu and Song, Xia},
  journal={arXiv preprint arXiv:2409.19913},
  year={2024}
}

@inproceedings{nawrot2023efficient,
  title={Efficient transformers with dynamic token pooling},
  author={Nawrot, Piotr and Chorowski, Jan and Lancucki, Adrian and Ponti, Edoardo Maria},
  booktitle={Proceedings of the 61st Annual Meeting of the Association for Computational Linguistics (Volume 1: Long Papers)},
  pages={6403--6417},
  year={2023}
}

@article{vaswani2017attention,
  title={Attention is all you need},
  author={Vaswani, Ashish and Shazeer, Noam and Parmar, Niki and Uszkoreit, Jakob and Jones, Llion and Gomez, Aidan N and Kaiser, {\L}ukasz and Polosukhin, Illia},
  journal={Advances in neural information processing systems},
  volume={30},
  year={2017}
}

@inproceedings{pepe2024taxonomy,
  title={A taxonomy of self-admitted technical debt in deep learning systems},
  author={Pepe, Federica and Zampetti, Fiorella and Mastropaolo, Antonio and Bavota, Gabriele and Di Penta, Massimiliano},
  booktitle={2024 IEEE International Conference on Software Maintenance and Evolution (ICSME)},
  pages={388--399},
  year={2024},
  organization={IEEE}
}

@inproceedings{pagnoni2025byte,
  title={Byte latent transformer: Patches scale better than tokens},
  author={Pagnoni, Artidoro and Pasunuru, Ramakanth and Rodriguez, Pedro and Nguyen, John and Muller, Benjamin and Li, Margaret and Zhou, Chunting and Yu, Lili and Weston, Jason E and Zettlemoyer, Luke and others},
  booktitle={Proceedings of the 63rd Annual Meeting of the Association for Computational Linguistics (Volume 1: Long Papers)},
  pages={9238--9258},
  year={2025}
}

@inproceedings{zouhar2023formal,
  title={A formal perspective on byte-pair encoding},
  author={Zouhar, Vil{\'e}m and Meister, Clara and Gastaldi, Juan and Du, Li and Vieira, Tim and Sachan, Mrinmaya and Cotterell, Ryan},
  booktitle={Findings of the Association for Computational Linguistics: ACL 2023},
  pages={598--614},
  year={2023}
}

@inproceedings{balestriero2022aug,
 author = {Balestriero, Randall and Misra, Ishan and LeCun, Yann},
 booktitle = {Advances in Neural Information Processing Systems},
 editor = {S. Koyejo and S. Mohamed and A. Agarwal and D. Belgrave and K. Cho and A. Oh},
 pages = {19631--19644},
 publisher = {Curran Associates, Inc.},
 title = {A Data-Augmentation Is Worth A Thousand Samples: Analytical Moments And Sampling-Free Training},
 url = {https://proceedings.neurips.cc/paper_files/paper/2022/file/7c080cab957edab671ac49ae11e51337-Paper-Conference.pdf},
 volume = {35},
 year = {2022}
}

@article{image-augs-survey,
	author = {Shorten, Connor and Khoshgoftaar, Taghi M.},
	date = {2019/07/06},
	doi = {10.1186/s40537-019-0197-0},
	id = {Shorten2019},
	isbn = {2196-1115},
	journal = {Journal of Big Data},
	number = {1},
	pages = {60},
	title = {A survey on Image Data Augmentation for Deep Learning},
	url = {https://doi.org/10.1186/s40537-019-0197-0},
	volume = {6},
	year = {2019}}

@article{li2023chatdoctor,
  title={Chatdoctor: A medical chat model fine-tuned on a large language model meta-ai (llama) using medical domain knowledge},
  author={Li, Yunxiang and Li, Zihan and Zhang, Kai and Dan, Ruilong and Jiang, Steve and Zhang, You},
  journal={Cureus},
  volume={15},
  number={6},
  year={2023},
  publisher={Cureus}
}

@misc{alpaca,
  author = {Rohan Taori and Ishaan Gulrajani and Tianyi Zhang and Yann Dubois and Xuechen Li and Carlos Guestrin and Percy Liang and Tatsunori B. Hashimoto },
  title = {Stanford Alpaca: An Instruction-following LLaMA model},
  year = {2023},
  publisher = {GitHub},
  journal = {GitHub repository},
  howpublished = {\url{https://github.com/tatsu-lab/stanford_alpaca}},
}

@article{shazeer2020glu,
  title={Glu variants improve transformer},
  author={Shazeer, Noam},
  journal={arXiv preprint arXiv:2002.05202},
  year={2020}
}

@article{yang2022image,
  title={Image data augmentation for deep learning: A survey},
  author={Yang, Suorong and Xiao, Weikang and Zhang, Mengchen and Guo, Suhan and Zhao, Jian and Shen, Furao},
  journal={arXiv preprint arXiv:2204.08610},
  year={2022}
}

@article{hoffmann2022training,
  title={Training compute-optimal large language models},
  author={Hoffmann, Jordan and Borgeaud, Sebastian and Mensch, Arthur and Buchatskaya, Elena and Cai, Trevor and Rutherford, Eliza and Casas, Diego de Las and Hendricks, Lisa Anne and Welbl, Johannes and Clark, Aidan and others},
  journal={arXiv preprint arXiv:2203.15556},
  year={2022}
}

@article{villalobos2022will,
  title={Will we run out of data? Limits of LLM scaling based on human-generated data},
  author={Villalobos, Pablo and Ho, Anson and Sevilla, Jaime and Besiroglu, Tamay and Heim, Lennart and Hobbhahn, Marius},
  journal={arXiv preprint arXiv:2211.04325},
  year={2022}
}

@article{radford2019language,
  title={Language models are unsupervised multitask learners},
  author={Radford, Alec and Wu, Jeffrey and Child, Rewon and Luan, David and Amodei, Dario and Sutskever, Ilya and others},
  journal={OpenAI blog},
  volume={1},
  number={8},
  pages={9},
  year={2019}
}

@article{prabhudesai2025diffusion,
  title={Diffusion beats autoregressive in data-constrained settings},
  author={Prabhudesai, Mihir and Wu, Mengning and Zadeh, Amir and Fragkiadaki, Katerina and Pathak, Deepak},
  journal={arXiv preprint arXiv:2507.15857},
  year={2025}
}

@inproceedings{zouhar2023tokenization,
  title={Tokenization and the noiseless channel},
  author={Zouhar, Vil{\'e}m and Meister, Clara and Gastaldi, Juan and Du, Li and Sachan, Mrinmaya and Cotterell, Ryan},
  booktitle={Proceedings of the 61st Annual Meeting of the Association for Computational Linguistics (Volume 1: Long Papers)},
  pages={5184--5207},
  year={2023}
}

@article{edmonds2002near,
  title={Near-synonymy and lexical choice},
  author={Edmonds, Philip and Hirst, Graeme},
  journal={Computational linguistics},
  volume={28},
  number={2},
  pages={105--144},
  year={2002},
  publisher={MIT Press One Rogers Street, Cambridge, MA 02142-1209, USA journals-info~…}
}

@article{team2025kimi,
  title={Kimi k2: Open agentic intelligence},
  author={Kimi Team and Bai, Yifan and Bao, Yiping and Chen, Guanduo and Chen, Jiahao and Chen, Ningxin and Chen, Ruijue and Chen, Yanru and Chen, Yuankun and Chen, Yutian and others},
  journal={arXiv preprint arXiv:2507.20534},
  year={2025}
}

@article{wang2024tokenization,
  title={Tokenization matters! degrading large language models through challenging their tokenization},
  author={Wang, Dixuan and Li, Yanda and Jiang, Junyuan and Ding, Zepeng and Luo, Ziqin and Jiang, Guochao and Liang, Jiaqing and Yang, Deqing},
  journal={arXiv preprint arXiv:2405.17067},
  year={2024}
}

@article{reddy2025much,
  title={How much is enough? the diminishing returns of tokenization training data},
  author={Reddy, Varshini and Schmidt, Craig W and Pinter, Yuval and Tanner, Chris},
  journal={arXiv preprint arXiv:2502.20273},
  year={2025}
}

@article{pimentel2020speakers,
  title={Speakers fill lexical semantic gaps with context},
  author={Pimentel, Tiago and Maudslay, Rowan Hall and Blasi, Damian and Cotterell, Ryan},
  journal={arXiv preprint arXiv:2010.02172},
  year={2020}
}

@inproceedings{chai-etal-2024-tokenization,
    title = "Tokenization Falling Short: On Subword Robustness in Large Language Models",
    author = "Chai, Yekun  and
      Fang, Yewei  and
      Peng, Qiwei  and
      Li, Xuhong",
    editor = "Al-Onaizan, Yaser  and
      Bansal, Mohit  and
      Chen, Yun-Nung",
    booktitle = "Findings of the Association for Computational Linguistics: EMNLP 2024",
    month = nov,
    year = "2024",
    address = "Miami, Florida, USA",
    publisher = "Association for Computational Linguistics",
    url = "https://aclanthology.org/2024.findings-emnlp.86/",
    doi = "10.18653/v1/2024.findings-emnlp.86",
    pages = "1582--1599"
}

@article{han2024parameter,
  title={Parameter-efficient fine-tuning for large models: A comprehensive survey},
  author={Han, Zeyu and Gao, Chao and Liu, Jinyang and Zhang, Jeff and Zhang, Sai Qian},
  journal={arXiv preprint arXiv:2403.14608},
  year={2024}
}

@article{chen2021regionvit,
  title={Regionvit: Regional-to-local attention for vision transformers},
  author={Chen, Chun-Fu and Panda, Rameswar and Fan, Quanfu},
  journal={arXiv preprint arXiv:2106.02689},
  year={2021}
}

@article{wang2023crossformer,
  title={Crossformer++: A versatile vision transformer hinging on cross-scale attention},
  author={Wang, Wenxiao and Chen, Wei and Qiu, Qibo and Chen, Long and Wu, Boxi and Lin, Binbin and He, Xiaofei and Liu, Wei},
  journal={IEEE Transactions on Pattern Analysis and Machine Intelligence},
  volume={46},
  number={5},
  pages={3123--3136},
  year={2023},
  publisher={IEEE}
}

@inproceedings{xu2021co,
  title={Co-scale conv-attentional image transformers},
  author={Xu, Weijian and Xu, Yifan and Chang, Tyler and Tu, Zhuowen},
  booktitle={Proceedings of the IEEE/CVF international conference on computer vision},
  pages={9981--9990},
  year={2021}
}

@article{chu2021twins,
  title={Twins: Revisiting the design of spatial attention in vision transformers},
  author={Chu, Xiangxiang and Tian, Zhi and Wang, Yuqing and Zhang, Bo and Ren, Haibing and Wei, Xiaolin and Xia, Huaxia and Shen, Chunhua},
  journal={Advances in neural information processing systems},
  volume={34},
  pages={9355--9366},
  year={2021}
}

@inproceedings{wang2021pyramid,
  title={Pyramid vision transformer: A versatile backbone for dense prediction without convolutions},
  author={Wang, Wenhai and Xie, Enze and Li, Xiang and Fan, Deng-Ping and Song, Kaitao and Liang, Ding and Lu, Tong and Luo, Ping and Shao, Ling},
  booktitle={Proceedings of the IEEE/CVF international conference on computer vision},
  pages={568--578},
  year={2021}
}

@inproceedings{liu2021swin,
  title={Swin transformer: Hierarchical vision transformer using shifted windows},
  author={Liu, Ze and Lin, Yutong and Cao, Yue and Hu, Han and Wei, Yixuan and Zhang, Zheng and Lin, Stephen and Guo, Baining},
  booktitle={Proceedings of the IEEE/CVF international conference on computer vision},
  pages={10012--10022},
  year={2021}
}

@misc{muennighoff2025scalingdataconstrainedlanguagemodels,
      title={Scaling Data-Constrained Language Models}, 
      author={Niklas Muennighoff and Alexander M. Rush and Boaz Barak and Teven Le Scao and Aleksandra Piktus and Nouamane Tazi and Sampo Pyysalo and Thomas Wolf and Colin Raffel},
      year={2025},
      eprint={2305.16264},
      archivePrefix={arXiv},
      primaryClass={cs.CL},
      url={https://arxiv.org/abs/2305.16264}, 
}

@inproceedings{wu-etal-2021-hi,
	title        = {Hi-Transformer: Hierarchical Interactive Transformer for Efficient and Effective Long Document Modeling},
	author       = {Wu, Chuhan  and Wu, Fangzhao  and Qi, Tao  and Huang, Yongfeng},
	year         = 2021,
	month        = aug,
	booktitle    = {Proceedings of the 59th Annual Meeting of the Association for Computational Linguistics and the 11th International Joint Conference on Natural Language Processing (Volume 2: Short Papers)},
	pages        = {848--853},
	doi          = {10.18653/v1/2021.acl-short.107},
	url          = {https://aclanthology.org/2021.acl-short.107/},
	editor       = {Zong, Chengqing  and Xia, Fei  and Li, Wenjie  and Navigli, Roberto}
}

@article{kudo2018subword,
  title={Subword regularization: Improving neural network translation models with multiple subword candidates},
  author={Kudo, Taku},
  journal={arXiv preprint arXiv:1804.10959},
  year={2018}
}

@article{jabbar2023morphpiece,
  title={Morphpiece: A linguistic tokenizer for large language models},
  author={Jabbar, Haris},
  journal={arXiv preprint arXiv:2307.07262},
  year={2023}
}

@inproceedings{agrawal2024translation,
  title={Translation errors significantly impact low-resource languages in cross-lingual learning},
  author={Agrawal, Ashish and Fazili, Barah and Jyothi, Preethi},
  booktitle={Proceedings of the 18th Conference of the European Chapter of the Association for Computational Linguistics (Volume 2: Short Papers)},
  pages={319--329},
  year={2024}
}

@inproceedings{miller-1994-wordnet,
    title = "{W}ord{N}et: A Lexical Database for {E}nglish",
    author = "Miller, George A.",
    booktitle = "{H}uman {L}anguage {T}echnology: Proceedings of a Workshop held at {P}lainsboro, {N}ew {J}ersey, {M}arch 8-11, 1994",
    year = "1994",
    url = "https://aclanthology.org/H94-1111/"
}

@inproceedings{lee2019set,
  title={Set transformer: A framework for attention-based permutation-invariant neural networks},
  author={Lee, Juho and Lee, Yoonho and Kim, Jungtaek and Kosiorek, Adam and Choi, Seungjin and Teh, Yee Whye},
  booktitle={International conference on machine learning},
  pages={3744--3753},
  year={2019},
  organization={PMLR}
}

@inproceedings{provilkov2020bpe,
  title={BPE-dropout: Simple and effective subword regularization},
  author={Provilkov, Ivan and Emelianenko, Dmitrii and Voita, Elena},
  booktitle={Proceedings of the 58th Annual Meeting of the Association for Computational Linguistics},
  pages={1882--1892},
  year={2020}
}

@inproceedings{unfairness-tokenizer,
 author = {Petrov, Aleksandar and La Malfa, Emanuele and Torr, Philip and Bibi, Adel},
 booktitle = {Advances in Neural Information Processing Systems},
 editor = {A. Oh and T. Naumann and A. Globerson and K. Saenko and M. Hardt and S. Levine},
 pages = {36963--36990},
 publisher = {Curran Associates, Inc.},
 title = {Language Model Tokenizers Introduce Unfairness Between Languages},
 url = {https://proceedings.neurips.cc/paper_files/paper/2023/file/74bb24dca8334adce292883b4b651eda-Paper-Conference.pdf},
 volume = {36},
 year = {2023}
}

@article{tokenization-meaning,
    author = {Haslett, David A.},
    title = {Tokenization Changes Meaning in Large Language Models: Evidence from Chinese},
    journal = {Computational Linguistics},
    volume = {51},
    number = {3},
    pages = {785-814},
    year = {2025},
    month = {09},
    issn = {0891-2017},
    doi = {10.1162/coli_a_00557},
    url = {https://doi.org/10.1162/coli_a_00557},
    eprint = {https://direct.mit.edu/coli/article-pdf/51/3/785/2507358/coli_a_00557.pdf},
}

@article{limisiewicz2023tokenization,
  title={Tokenization impacts multilingual language modeling: Assessing vocabulary allocation and overlap across languages},
  author={Limisiewicz, Tomasz and Balhar, Ji{\v{r}}{\'\i} and Mare{\v{c}}ek, David},
  journal={arXiv preprint arXiv:2305.17179},
  year={2023}
}

@misc{zheng2025brokentokenslanguagemodel,
      title={Broken Tokens? Your Language Model can Secretly Handle Non-Canonical Tokenizations}, 
      author={Brian Siyuan Zheng and Alisa Liu and Orevaoghene Ahia and Jonathan Hayase and Yejin Choi and Noah A. Smith},
      year={2025},
      eprint={2506.19004},
      archivePrefix={arXiv},
      primaryClass={cs.CL},
      url={https://arxiv.org/abs/2506.19004}, 
}

@inproceedings{cosma-etal-2025-strawberry,
    title = "The Strawberry Problem: Emergence of Character-level Understanding in Tokenized Language Models",
    author = "Cosma, Adrian  and
      Ruseti, Stefan  and
      Radoi, Emilian  and
      Dascalu, Mihai",
    editor = "Christodoulopoulos, Christos  and
      Chakraborty, Tanmoy  and
      Rose, Carolyn  and
      Peng, Violet",
    booktitle = "Proceedings of the 2025 Conference on Empirical Methods in Natural Language Processing",
    month = nov,
    year = "2025",
    address = "Suzhou, China",
    publisher = "Association for Computational Linguistics",
    url = "https://aclanthology.org/2025.emnlp-main.1434/",
    doi = "10.18653/v1/2025.emnlp-main.1434",
    pages = "28240--28251",
    ISBN = "979-8-89176-332-6"
}

@inproceedings{whittington2025tokenisation,
  title={Tokenisation is np-complete},
  author={Whittington, Philip and Bachmann, Gregor and Pimentel, Tiago},
  booktitle={Proceedings of the 63rd Annual Meeting of the Association for Computational Linguistics (Volume 1: Long Papers)},
  pages={28133--28153},
  year={2025}
}

@inproceedings{deiseroth-etal-2024-free,
	title        = {{T}-{FREE}: Subword Tokenizer-Free Generative {LLM}s via Sparse Representations for Memory-Efficient Embeddings},
	author       = {Deiseroth, Bj{\"o}rn  and Brack, Manuel  and Schramowski, Patrick  and Kersting, Kristian  and Weinbach, Samuel},
	year         = 2024,
	month        = nov,
	booktitle    = {Proceedings of the 2024 Conference on Empirical Methods in Natural Language Processing},
	publisher    = {Association for Computational Linguistics},
	address      = {Miami, Florida, USA},
	pages        = {21829--21851},
	doi          = {10.18653/v1/2024.emnlp-main.1217},
	url          = {https://aclanthology.org/2024.emnlp-main.1217/},
	editor       = {Al-Onaizan, Yaser  and Bansal, Mohit  and Chen, Yun-Nung}
}

@inproceedings{he2024hdt,
	title        = {{HDT}: Hierarchical Document Transformer},
	author       = {Haoyu He and Markus Flicke and Jan Buchmann and Iryna Gurevych and Andreas Geiger},
	year         = 2024,
	booktitle    = {First Conference on Language Modeling},
	url          = {https://openreview.net/forum?id=dkpeWQRmlc}
}

@article{chalkidis2022exploration,
	title        = {An exploration of hierarchical attention transformers for efficient long document classification},
	author       = {Chalkidis, Ilias and Dai, Xiang and Fergadiotis, Manos and Malakasiotis, Prodromos and Elliott, Desmond},
	year         = 2022,
	journal      = {arXiv preprint arXiv:2210.05529}
}

@inproceedings{chen2021crossvit,
	title        = {Crossvit: Cross-attention multi-scale vision transformer for image classification},
	author       = {Chen, Chun-Fu Richard and Fan, Quanfu and Panda, Rameswar},
	year         = 2021,
	booktitle    = {Proceedings of the IEEE/CVF international conference on computer vision},
	pages        = {357--366}
}

@inproceedings{harzig2022synchronized,
  title={Synchronized audio-visual frames with fractional positional encoding for transformers in video-to-text translation},
  author={Harzig, Philipp and Einfalt, Moritz and Lienhart, Rainer},
  booktitle={2022 IEEE International Conference on Image Processing (ICIP)},
  pages={2041--2045},
  year={2022},
  organization={IEEE}
}

@InProceedings{gomez2024depression,
author="Gimeno-G{\'o}mez, David
and Bucur, Ana-Maria
and Cosma, Adrian
and Mart{\'i}nez-Hinarejos, Carlos-David
and Rosso, Paolo",
editor="Goharian, Nazli
and Tonellotto, Nicola
and He, Yulan
and Lipani, Aldo
and McDonald, Graham
and Macdonald, Craig
and Ounis, Iadh",
title="Reading Between the Frames: Multi-modal Depression Detection in Videos from Non-verbal Cues",
booktitle="Advances in Information Retrieval",
year="2024",
publisher="Springer Nature Switzerland",
address="Cham",
pages="191--209",
isbn="978-3-031-56027-9"
}

@inproceedings{singh2024aya,
  title={Aya dataset: An open-access collection for multilingual instruction tuning},
  author={Singh, Shivalika and Vargus, Freddie and D’souza, Daniel and Karlsson, B{\"o}rje F and Mahendiran, Abinaya and Ko, Wei-Yin and Shandilya, Herumb and Patel, Jay and Mataciunas, Deividas and O’Mahony, Laura and others},
  booktitle={Proceedings of the 62nd Annual Meeting of the Association for Computational Linguistics (Volume 1: Long Papers)},
  pages={11521--11567},
  year={2024}
}

@misc{aryabumi2024aya,
      title={Aya 23: Open Weight Releases to Further Multilingual Progress}, 
      author={Viraat Aryabumi and John Dang and Dwarak Talupuru and Saurabh Dash and David Cairuz and Hangyu Lin and Bharat Venkitesh and Madeline Smith and Kelly Marchisio and Sebastian Ruder and Acyr Locatelli and Julia Kreutzer and Nick Frosst and Phil Blunsom and Marzieh Fadaee and Ahmet Üstün and Sara Hooker},
      year={2024},
      eprint={2405.15032},
      archivePrefix={arXiv},
      primaryClass={cs.CL}
}

@inproceedings{shin2024large,
	title        = {Large Language Models Lack Understanding of Character Composition of Words},
	author       = {Andrew Shin and Kunitake Kaneko},
	year         = 2024,
	booktitle    = {ICML 2024 Workshop on LLMs and Cognition},
	url          = {https://openreview.net/forum?id=oP5FXcPAeG}
}

@inproceedings{mcleish2024abacus,
	title        = {Transformers Can Do Arithmetic with the Right Embeddings},
	author       = {Sean Michael McLeish and Arpit Bansal and Alex Stein and Neel Jain and John Kirchenbauer and Brian R. Bartoldson and Bhavya Kailkhura and Abhinav Bhatele and others},
	year         = 2024,
	booktitle    = {The Thirty-eighth Annual Conference on Neural Information Processing Systems},
	url          = {https://openreview.net/forum?id=aIyNLWXuDO}
}

\appendix
\section{Appendix}
\label{sec:appendix}
\subsection{Discarded Alternative Designs}
\label{sec:alt-designs}

Several alternative ways of incorporating alternative tokenizations were considered and discarded due to practical or conceptual limitations.

\paragraph{Direct stochastic tokenization for next-token prediction.}
\citet{chai-etal-2024-tokenization} use BPE-dropout in encoder–decoder machine translation, where the source and target play asymmetric roles and there is no causal label conflict. In contrast, we operate in a pure language modeling setting with next-token prediction. If the input stream itself were tokenized stochastically at training time, the notion of the \textit{correct} next token becomes under-specified: should the label be the canonical token, one of many valid stochastic segmentations, or some mixture? This would require redefining the loss and the evaluation protocol. We instead keep the canonical interface as the only label space and treat subword variants as auxiliary, unlabeled augmentations.

\paragraph{Training directly on subtokens only.}
One possibility is to abandon canonical tokens entirely and train on subtoken sequences alone. However, this again raises the question of what constitutes the target: should the model predict the next subtoken, the completion of a sampled segmentation, or a canonical segmentation reconstructed from subtokens? Each choice bakes in a particular inductive bias about the "true" tokenization and complicates both the objective and deployment. Our design avoids this by keeping the main model and its loss completely standard (canonical next-token prediction) and using subword variants only as additional context.

\paragraph{Multi-task prediction of subtoken structure.}
Another alternative is to add an auxiliary task that predicts the subword variants decomposition of each canonical token (e.g., predicting $s_t$ from $x_t$). This introduces additional labels and multi-task optimization challenges: the model must simultaneously learn language modeling and tokenizer reconstruction, potentially leading to conflicting gradients and unclear trade-offs. Moreover, the label space for the auxiliary task is itself ambiguous whenever multiple segmentations are valid. In contrast, our cross-attention-based design only conditions on sampled subword variants; it does not predict them, and thus avoids auxiliary supervision.

\paragraph{Single-branch decoder with concatenated tokens and subtokens.}
One could concatenate canonical tokens and subword variants into a single long sequence and feed it to a larger causal decoder. This provides less control over how and where subword variants information enters the computation: every layer must process a longer sequence, increasing memory and compute without an explicit mechanism for aligning subtokens to their canonical parents. Our two-branch design isolates the augmentations in a small causal encoder and injects them only where needed, at a chosen depth in the main trunk.

\paragraph{Additive fusion of subword variants embeddings.}
A simpler fusion strategy would be to aggregate the subword variants embeddings belonging to token $s_t$ (e.g., by averaging) and add them to the embedding of $s_t$. While cheap, this forces all subtokens to contribute uniformly and removes the model's ability to focus on task-relevant segments or ignore noisy ones. Cross-attention, by contrast, lets the model learn task-dependent weights over the subword variants representations, deciding for each context how much to use or disregard each subtoken.

\subsection{Hyperparameters}
\label{sec:hyperparams}
The standard transformer has 8 blocks with $d_{model} = 512 \times\mu$, parametrized by a width multiplier $\mu$ \cite{yang2022tensorprogramsvtuning}, using SwiGLU \cite{shazeer2020glu} and RMSNorm \cite{zhang2019root}. Our variant has a separate, one-block causal encoder, having a fixed $d_{model}^{enc} = 256$. We used a context size of 512 tokens for both models. We tested two model sizes, corresponding to $\mu \in \{1, 2\}$. The models have the number of parameters $N_{params}(\mu = 1) = 88M$ and $N_{params}(\mu = 2) = 244M$, when using the GPT-2 tokenizer vocabulary. The causal encoder has a fixed number of 1M parameters for all model sizes. Across experiments, we used AdamW \cite{kingma2014adam} optimizer, with $\beta_1 = 0.9$, $\beta_2 = 0.95$, $eps=10^{-10}$, $wd=0.01$, with a per-device batch size of 32, and a learning rate of 0.00001, annealed using a cosine decay scheduler \cite{loshchilov2016sgdr}. The learning rate was adjusted according to $\mu$P \cite{yang2022tensorprogramsvtuning}. We trained all runs on 4 NVIDIA A100 GPUs. A training run lasted between 4 hours for the smallest model and 24 hours for the largest. No dropout or attention dropout was used during standard pretraining runs. 

\noindent \textbf{Parameter count.} For the encoder, we have a $d_{enc} \cdot  \mu = d_{dec}$, with $\mu$ being the multiplicative factor for the model size. In practice, we use linear layers that function as adapters between the widths of the causal decoder and encoder. We have two such adapters: $L_{emb}: \mathbb{R}^{d_{dec}} \to \mathbb{R}^{d_{enc}}$ which translates the token embedding matrix into a smaller dimensionality, and $L_{aux}: \mathbb{R}^{d_{enc}} \to \mathbb{R}^{d_{dec}}$ which translates back the token embeddings to the dimensionality of the main trunk. 

For example, under the usual Transformer parameter approximation $P_{\mathrm{enc}}(d_{\mathrm{enc}}) \approx c_{\mathrm{enc}} d_{\mathrm{enc}}^2$, the auxiliary branch scales as $P_{\mathrm{aux}}(\mu) \approx (c_{\mathrm{enc}} + 2\mu)d_{\mathrm{enc}}^2.$ Moving from $\mu=1$ to $\mu=2$ therefore increases the auxiliary branch from $(c_{\mathrm{enc}}+2)d_{\mathrm{enc}}^2$ to $(c_{\mathrm{enc}}+4)d_{\mathrm{enc}}^2$, rather than doubling the entire auxiliary module. Therefore, the total auxiliary branch grows from roughly $1.05\mathrm{M}$ to $1.31\mathrm{M}$ parameters between the two model scales. The auxiliary branch is thus not strictly parameter-constant, but most of its parameters come from the fixed-width encoder core; only the adapter projections scale with the decoder width.

\subsection{Tokenizer statistics}
\label{sec:tok-stats}
In Table \ref{tab:tokenizer-stats} we show statistics for the GPT-2 and AYA tokenizer across the datasets we used in this work. 

\begin{table*}[hbt!]
    \centering
    \resizebox{\linewidth}{!}{
    \begin{tabular}{lllrrrrrr}
\textbf{Dataset} & \textbf{Lang.} & \textbf{Tokenizer} & \textbf{Canonical Seq. Len.} & \textbf{Aux Seq. Len.} & \textbf{Aux/Can Ratio} & \textbf{\% Unsplit Tokens} & \textbf{Tok/Word} & \textbf{Tok/Byte}\\
\midrule
\multirow[t]{2}{*}{\textbf{FineWeb-Edu}} & \multirow[t]{2}{*}{ } & aya-23 & 145.89 & 303.64 & 2.08 & 17.13 & 1.27 & 0.21 \\
 &  & GPT-2 & 148.28 & 320.81 & 2.16 & 15.05 & 1.29 & 0.21 \\
\cline{1-9} \cline{2-9}
\multirow[t]{2}{*}{\textbf{ChatDoctor}} & \multirow[t]{2}{*}{ } & aya-23 & 86.58 & 178.73 & 2.06 & 18.00 & 1.34 & 0.22 \\
 &  & {GPT-2} & 89.15 & 187.52 & 2.10 & 18.19 & 1.38 & 0.23 \\
\cline{1-9} \cline{2-9}
\multirow[t]{2}{*}{\textbf{C++}} & \multirow[t]{2}{*}{ } & aya-23 & 88.23 & 158.69 & 1.80 & 34.19 & 1.98 & 0.33 \\
 &  & GPT-2 & 93.07 & 167.09 & 1.80 & 35.68 & 2.09 & 0.35 \\
\cline{1-9} \cline{2-9}
\multirow[t]{2}{*}{\textbf{Haskell}} & \multirow[t]{2}{*}{ } & aya-23 & 184.80 & 357.61 & 1.94 & 25.05 & 1.58 & 0.26 \\
 &  & GPT-2 & 195.40 & 385.90 & 1.97 & 25.20 & 1.67 & 0.27 \\
\cline{1-9} \cline{2-9}
\multirow[t]{2}{*}{\textbf{AlpacaEval}} & \multirow[t]{2}{*}{ } & aya-23 & 76.20 & 150.84 & 1.98 & 24.22 & 1.31 & 0.22 \\
 &  & GPT-2 & 77.24 & 158.86 & 2.06 & 22.58 & 1.33 & 0.22 \\
\cline{1-9} \cline{2-9}
\multirow[t]{2}{*}{\textbf{ChatRestaurant}} & \multirow[t]{2}{*}{ } & aya-23 & 55.36 & 111.31 & 2.01 & 20.41 & 1.32 & 0.22 \\
 &  & GPT-2 & 57.89 & 120.20 & 2.08 & 19.54 & 1.39 & 0.23 \\
\midrule
\multirow[t]{28}{*}{\textbf{AYA-23}} & \multirow[t]{2}{*}{eng} & aya-23 & 100.28 & 202.97 & 2.02 & 21.48 & 1.36 & 0.22 \\
 &  & GPT-2 & 102.79 & 215.15 & 2.09 & 20.23 & 1.39 & 0.22 \\
\cline{2-9}
 & \multirow[t]{2}{*}{fil} & aya-23 & 116.98 & 236.59 & 2.02 & 19.56 & 1.91 & 0.32 \\
 &  & GPT-2 & 132.90 & 267.07 & 2.01 & 18.24 & 2.17 & 0.36 \\
\cline{2-9}
 & \multirow[t]{2}{*}{fin} & aya-23 & 189.36 & 389.46 & 2.06 & 15.22 & 3.32 & 0.37 \\
 &  & GPT-2 & 220.95 & 429.84 & 1.95 & 19.24 & 3.88 & 0.43 \\
\cline{2-9}
 & \multirow[t]{2}{*}{fra} & aya-23 & 117.14 & 239.94 & 2.05 & 21.25 & 1.53 & 0.24 \\
 &  & GPT-2 & 162.25 & 332.25 & 2.05 & 17.62 & 2.13 & 0.33 \\
\cline{2-9}
 & \multirow[t]{2}{*}{ind} & aya-23 & 123.70 & 255.06 & 2.06 & 21.82 & 1.68 & 0.23 \\
 &  & GPT-2 & 200.88 & 400.45 & 1.99 & 18.25 & 2.72 & 0.38 \\
\cline{2-9}
 & \multirow[t]{2}{*}{nld} & aya-23 & 114.60 & 233.00 & 2.03 & 22.84 & 1.60 & 0.26 \\
 &  & GPT-2 & 162.35 & 325.16 & 2.00 & 18.38 & 2.27 & 0.37 \\
\cline{2-9}
 & \multirow[t]{2}{*}{nya} & aya-23 & 170.20 & 342.98 & 2.02 & 18.89 & 2.64 & 0.37 \\
 &  & GPT-2 & 194.43 & 378.90 & 1.95 & 19.30 & 3.01 & 0.42 \\
\cline{2-9}
 & \multirow[t]{2}{*}{plt} & aya-23 & 99.70 & 196.91 & 1.98 & 22.20 & 2.48 & 0.36 \\
 &  & GPT-2 & 114.93 & 217.76 & 1.89 & 25.17 & 2.86 & 0.42 \\
\cline{2-9}
 & \multirow[t]{2}{*}{por} & aya-23 & 67.99 & 135.58 & 1.99 & 25.61 & 1.46 & 0.24 \\
 &  & GPT-2 & 101.33 & 200.03 & 1.97 & 22.00 & 2.18 & 0.35 \\
\cline{2-9}
 & \multirow[t]{2}{*}{sna} & aya-23 & 140.96 & 283.78 & 2.01 & 18.83 & 2.82 & 0.35 \\
 &  & GPT-2 & 161.57 & 314.33 & 1.95 & 20.61 & 3.24 & 0.40 \\
\cline{2-9}
 & \multirow[t]{2}{*}{spa} & aya-23 & 92.58 & 190.23 & 2.05 & 21.50 & 1.41 & 0.23 \\
 &  & GPT-2 & 136.95 & 277.62 & 2.03 & 19.43 & 2.09 & 0.34 \\
\cline{2-9}
 & \multirow[t]{2}{*}{swe} & aya-23 & 132.70 & 262.70 & 1.98 & 24.16 & 2.19 & 0.33 \\
 &  & GPT-2 & 149.76 & 299.15 & 2.00 & 17.58 & 2.47 & 0.37 \\
\cline{2-9}
 & \multirow[t]{2}{*}{zsm} & aya-23 & 94.53 & 196.88 & 2.08 & 19.47 & 1.73 & 0.24 \\
 &  & GPT-2 & 149.87 & 298.19 & 1.99 & 18.22 & 2.74 & 0.39 \\
\cline{2-9}
 & \multirow[t]{2}{*}{zul} & aya-23 & 234.76 & 470.74 & 2.01 & 18.50 & 3.33 & 0.39 \\
 &  & GPT-2 & 273.16 & 511.51 & 1.87 & 26.14 & 3.87 & 0.45 \\
\end{tabular}}
    \caption{Descriptive statistics for canonical (Can.) and auxiliary (Aux.) tokenizations across datasets, languages, and tokenizers. These statistics quantify both the additional sequence-level overhead introduced by the auxiliary branch and the amount of tokenization variation available to the method in each setting.}
    \label{tab:tokenizer-stats}
\end{table*}

\end{document}